\def\year{2021}\relax
\documentclass[letterpaper]{article} % DO NOT CHANGE THIS
\usepackage{aaai21}  % DO NOT CHANGE THIS
\usepackage{times}  % DO NOT CHANGE THIS
\usepackage{helvet} % DO NOT CHANGE THIS
\usepackage{courier}  % DO NOT CHANGE THIS
\usepackage[hyphens]{url}  % DO NOT CHANGE THIS
\usepackage{graphicx} % DO NOT CHANGE THIS
\usepackage{graphicx}  % DO NOT CHANGE THIS
\usepackage{natbib}  % DO NOT CHANGE THIS AND DO NOT ADD ANY OPTIONS TO IT
\usepackage{caption} % DO NOT CHANGE THIS AND DO NOT ADD ANY OPTIONS TO IT
\usepackage{amsmath}
\usepackage{amssymb}
\usepackage{amsthm}
\title{Functional Neural Network (FNN) for decision processing, a racing network of programmable neurons where the operating model is the network itself}
\title{Functional neural network for decision processing, a racing network of programmable neurons with fuzzy logic where the target operating model relies on the network itself}
\author {
        Frederic Jumelle,\textsuperscript{\rm 1}
        Kelvin So,\textsuperscript{\rm 1}
        Didan Deng\textsuperscript{\rm 2} \\
}
\affiliations {
    \textsuperscript{\rm 1} Bright Nation Limited, Smart-Space 3F, Cyberport, Hong Kong\\
    \textsuperscript{\rm 2} Neuromorphic Interactive System Laboratory, Department of Electronic and Computer Engineering, The Hong Kong University of Science and Technology, Hong Kong \\
    f.jumelle@brightnationlimited.com, ccso@brightnationlimited.com, ddeng@ust.hk
}
\begin{document}

\maketitle

\begin{abstract}
In this paper, we are introducing a novel model of artificial intelligence, the functional neural network for modeling of human decision-making processes. This neural network is composed of multiple artificial neurons racing in the network. Each of these neurons has a similar structure programmed independently by the users and composed of an intention wheel, a motor core and a sensory core representing the user itself and racing at a specific velocity. The mathematics of the neuron's formulation and the racing mechanism of multiple nodes in the network will be discussed, and the group decision process with fuzzy logic and the transformation of these conceptual methods into practical methods of simulation and in operations will be developed. Eventually, we will describe some possible future research directions in the fields of finance, education and medicine including the opportunity to design an intelligent learning agent with application in business operations supervision. We believe that this functional neural network has a promising potential to transform the way we can compute decision-making and lead to a new generation of neuromorphic chips for seamless human-machine interactions.
\end{abstract}

\section{Introduction}
Modeling the internal dynamics of human central nervous system has been a prominent and significant research problem in artificial intelligence (AI) for decades. Not only is such knowledge appealing for intellectual purpose, but is also important for the advance of brain-inspired AI technology and strong AI, i.e. machines that can experience consciousness and aim to achieve human cognitive abilities~\cite{long2019review, alpcan2017toward, vsekrst2020ai}. However, the major streams of AI technology focus on weak AI, i.e. the use of machines for analysing and accomplishing well-defined and specific problem solving and reasoning tasks with certain given rules~\cite{gams1997weak, rajan2017towards}. Examples of weak AI include the most common applications of AI, namely, facial recognition, which analyse human facial image in a pixel scale and is mostly based on the framework of convolutional neural network (CNN) to extract features of human faces with trained filters~\cite{sun2018face, mahmood2017review}. Competency of these models of weak AI is further improved by enhancement of computational efficiency than by comprehensive and rigorous understanding of how intelligence is generated or processed in human central nervous system~\cite{strong2016applications}. 

As a recent breakthrough in deep learning, generative adversarial network (GAN) suggests a new direction on strong AI research. GAN comprises of two alternating neural networks, the generator and the discriminator, and attempts to mimic the computing power of the human brain in term of learning to distinguish real data from fake data~\cite{goodfellow2016nips, goodfellow2014generative}. However, instead of being inspired from the neuro-biological architectures and transmission mechanisms of the central nervous system, the intrinsic architecture of GAN is based on the CNN framework, making it less appealing for modelling the decision-making process of humans and brings scarce insight into understanding the mechanism of human central nervous system. On the other side, Spiking Neural Networks (SNN) are networks that more closely mimic natural brain networks~\cite{ghosh2009spiking, wang2020supervised}. In addition to neuronal and synaptic state, SNNs incorporate the concept of time into the operating model. The idea is that neurons in the SNN do not fire at each propagation cycle but fire only when the neuron’s electrical charge reaches a specific value.

Decision-making is a complex process with multiple inputs, a fuzzy logic and an output unique to a person, which requires a computing model capable of operating a timed competition of neurons programmed upon personal attributes such as neuropsychological performance and central nervous system processing latency. This neuromorphic approach has been put in place wherein the neurons are programmed by the users to create a racing network for decision processing where the operating model is the network itself. This model refers to the function(s) of learning by accumulating try and succeed or try and fail processes such as those driven by the urge to execute and those by the urge to engage, along with the long short term memory of them. One of the assumptions is that decision making involves at least 4 components, namely the intention, the reaction time, the emotional response and the velocity of the process, that can be modelled in a way that computing them becomes possible. The second assumption is that decision-making is a learning process for which reinforcement is a key component based on success or failure when the record of these outputs can be kept in memory. The third assumption is that human logic is fuzzy in which the truth values of variables may be any real number between totally true and totally false, between 0 and 1 both inclusive. However, in human society a decision is expected to be made in terms of a simplistic yes or no. This extreme bipolarity is quite challenging and can throw light onto what researchers have called the mirror neuron system in the learning process especially in action observation and action execution~\cite{shillcock2019mirror, zhang2018activation}, which could be playing a significant role in human decision making process such as in the case of moral dilemma~\cite{christov2017deontological}. This observation makes the invention of a novel neural network architecture with the insight of the human mirror neuron system~\cite{kilroy2017neuroimaging}, a keen step towards understanding the central nervous system and computing artificial intelligence. 

In this paper, we are proposing a novel artificial general intelligence model, the Functional Neural Network (FNN), which refers to the capacity of a neuron to become part of a neural network through temporal synchronization. FNN is made of programmable neurons based on a 3-dimensional profile (using attributes such as biological-metrics, neuropsychological-metrics and chrono-metrics) of each user generated from a neuropsychological performance test taking inputs from the user's mobile camera for facial attributes recognition and emotional state capture. The Artificial Mirror Neuron (AMN) is a single hybrid neuron created upon the user's profile scores which comprises a mirror structure articulated between a motor core and a sensory core, and determines the velocity of a user's neuron in a network of multiple users represented by other neurons in a race. This competition is used for modelling the decision-making process of each user as seen in a group of other users of similar, competitive, or challenging nature which depends on the type of their individual request for decision processing. Decision types can be simple binary choice or reactional to sentiment analysis or a probabilistic response to a specific market signal whether social, industrial, or financial. Given sufficient training by a variety of users under a wide set of requests that are kept in memory, the FNN is able to generate a prescription to a user's question or dilemma, upon group consensus and/or expert consensus for the user's reference and trigger an executive decision. Group consensus and expert consensus are a breakthrough from existing artificial intelligence techniques and offer a new perspective in understanding the importance of synchronicity in human decision processing with potential application in industry sectors where rational decision-making matters greatly.

\section{Methodology}

\begin{figure*}[t]
\centering
\includegraphics[width=0.7\textwidth]{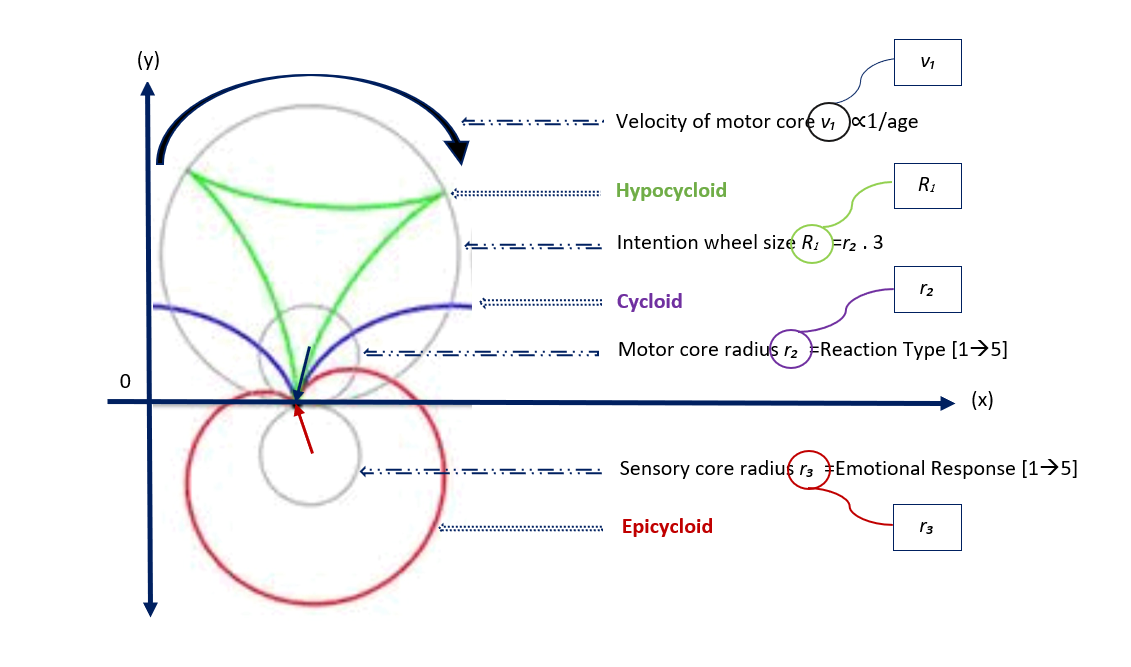} % Reduce the figure size so that it is slightly narrower than the column.
\caption{Tri-cycloid structure of the artificial mirror neuron.}
\label{1}
\end{figure*}

\subsection{Artificial Mirror Neuron (AMN) of the Functional Neural Network}
Artificial Mirror Neuron (AMN) is a hybrid structure serving as the fundamental building block of the Functional Neural Network (FNN). A single functional network comprises a large network of mirror neurons. Each of them is a programmable single unit which has multiple inputs. The inputs of each neuron are obtained at the time of launching a new request by using a simple cognitive test that can quantify the subject's reaction time, combined with the MIMAMO-net~\cite{deng2019mimamo} that can qualify the subject's emotional response by facial emotion recognition. In other words, FNN is an AI model based on a network of specialized neurons, i.e. AMN, to simulate some user brain functions, obtain a reaction and generate a prescription useful to the user. 

Each neuron is made of 3 basic elements: the intention wheel (large circle above the axis x with the green hypocycloid with 3 cusps), the motor core inside (small circle inside the intention wheel and above the axis x) and its counterpart the sensory core (small circle and below the axis x with the red epicycloid with one cusp). The neuron moves dynamically from left to right in a calculated velocity depending on the age of the user. 

A schematic structural view of an AMN is presented in Figure \ref{1}.
 
The 4 featured components of AMN includes:

\begin{itemize}
\item Size of intention wheel $R_1$: the intention wheel is a fictitious component of an intention-based decision since it is hardly measurable, it only refers to the initial driver of a decision for which the size is apprehended by a coefficient, usually 3 in reference to the triangle made of the occipital lobe, the frontal cortex and the thalamus, applied to the motor core; 

\item Radius of the motor core $r_2$: the motor core refers to the reaction time of the user in the programming test, which depends on the relative cognitive velocity of the subject in a range of 1 to 5. This score sets the radius of the motor core; 

\item Radius of the sensory core $r_3$: the sensory core refers to the emotional response of the user in the same test mirroring the reaction time, which depends on the valence-arousal status of the user in a range of 1 to 5. This score sets the radius of the sensory core:  

\item Velocity of the feed forward system $v_1$: the velocity of the feed forward system is the motor’s velocity and inversely proportional to the age of the user, i.e. the decision maker.
\end{itemize}

To initialize a mirror neuron, the inputs are:
 
 \begin{itemize}
\item the size of the intention wheel $R_1$
\item the motor core’s radius $r_1$
\item the sensory core’s radius $r_2$
\item the velocity of the motor core $v_1$
\end{itemize}

The internal dynamic of each Artificial Mirror Neuron is tri-cycloid whereas the intention wheel takes the form of a hypocycloid, the motor core describes a cycloid and the sensory core an epicycloid. When these 3 elements are combined and put in motion, they can compute the behavior’s complexity of a single unit of the natural architecture involved in functional decision processing. They can also be used to construct a model of acquisition of social skills such as inter-personal communication of information whether by imitation or reprocessing.

\subsection{Mathematical Formulations of AMN}  

Now we present the mathematical formulations of AMN and outline the formulas governing the dynamics of the intention wheel, motor core and sensory core. 

\subsubsection{Motor core}

The trajectory of the motor is a cycloid, which is the curve traced by a point of the rim of a circular wheel as the wheel rolls along a straight line without slipping.

The cycloid (motor core) runs through the origin, with a horizontal base given by the x-axis, generated by a circle of radius r rolling over the positive side of the base ($y \geq 0$). Its trajectory consists of the points $(x, y)$, with  
\begin{align}
x &= r \left( \theta - \sin \theta \right) \\
y &= r \left( 1 - \cos \theta \right)
\end{align}
where $\theta$ is a real parameter, corresponding to the angle through which the rolling circle has rotated. (See figure 3). For given $\theta$, the circle’s center lies at $(x, y) = (r\theta, r)$. 

Solving for $\theta$ and replacing, the Cartesian equation of the motor core is found to be: 
\begin{align}
x = r \cos ^{-1} \left( 1 - \frac{y}{r}\right) - \sqrt{y(2r-y)}
\end{align}

On the other hand, when $y$ is viewed as a function of $x$, the cycloid is differentiable everywhere except at the cusps, where it hits the $x$-axis, with the derivative tending toward $\inf$ or $-\inf$ as one approaches a cusp. The map from $\theta$ to $(x, y)$ is a differentiable curve, and the singularity where the derivative is $0$ is an ordinary cusp. 

A cycloid segment from one cusp to the next is called an arch of the cycloid. The first arch of the cycloid consists of points such that $0\leq \theta\leq 2 \pi$. All in all, the equation of the cycloid (motor core) satisfies the differential equation: 
\begin{align}
\left( \frac{dy}{dx} \right)^2 = \frac{2r}{y} - 1
\end{align}

\subsubsection{Sensory core}

Sensory core is an epicycloid, which is a plane curve produced by tracing the path of a chosen point on the circumference of a circle – called an epicycle – which rolls without slipping around a fixed circle.

If the small cycle has radius r, and the larger circle has radius R=kr, then the parametric equations for the trajectory of the epicycloid (sensory core) is given by either: 
\begin{align}
x &= \left( R+r \right) \cos \theta - r \cos \left( \frac{R+r}{r}\theta \right) \\
y &= \left( R+r \right) \sin \theta - r \sin \left( \frac{R+r}{r}\theta \right)
\end{align}
or
\begin{align}
x &= r \left( k+1 \right) \cos \theta - r \cos \left(  \left( k+1 \right) \theta \right) \\
y &= r \left( k+1 \right) \sin \theta - r \sin \left(  \left( k+1 \right) \theta \right)
\end{align}
where $\theta$ can be referred to in figure 3.

We can observe that if $k$ is an integer, then the curve is closed, and has $k$ cusps. If $k$ is a rational number, say $k=p/q$ expressed in simplest terms, then the curve has $p$ cusps. If $k$ is an irrational number, then the curve never closes, and forms a dense subset of the space between the larger circle and a circle of the radius $R+2r$.

\subsubsection{Intention wheel}

An intention wheel is a hypocycloid, which is a special plane curve generated by the trace of a fixed point on a small circle that rolls within a larger circle. In AMN, the hypocycloid of the intention wheel depends completely on the cycloid of the motor core. 

If the small cycle has radius r, and the larger circle has radius R=kr, then the parametric equations for the trajectory of the hypocycloid (intention wheel) is given by either: 
\begin{align}
x &= \left( R-r \right) \cos \theta + r \cos \left( \frac{R-r}{r}\theta \right) \\
y &= \left( R-r \right) \sin \theta - r \sin \left( \frac{R-r}{r}\theta \right)
\end{align}
or
\begin{align}
x &= r \left( k-1 \right) \cos \theta + r \cos \left(  \left( k-1 \right) \theta \right) \\
y &= r \left( k-1 \right) \sin \theta - r \sin \left(  \left( k-1 \right) \theta \right)
\end{align}

We can observe that if $k$ is an integer, then the curve is closed, and has $k$ cusps. If $k$ is a rational number, say $k=p/q$ expressed in simplest terms, then the curve has $p$ cusps. If $k$ is an irrational number, then the curve never closes, and fills the space between the larger circle and a circle of radius $R-2r$. 

Lastly, the area enclosed by the hypocycloid and the arc length of the hypocycloid are given by:
\begin{align}
A = \frac{(k-1)(k-2)}{k^2} \pi R^2 = (k-1)(k-2) \pi r^2
\end{align} 
and
\begin{align}
s = \frac{8(k-1)}{k} \pi R = 8(k-1)r
\end{align} 
respectively.

\begin{figure}[t]
\centering
\includegraphics[width=1\columnwidth]{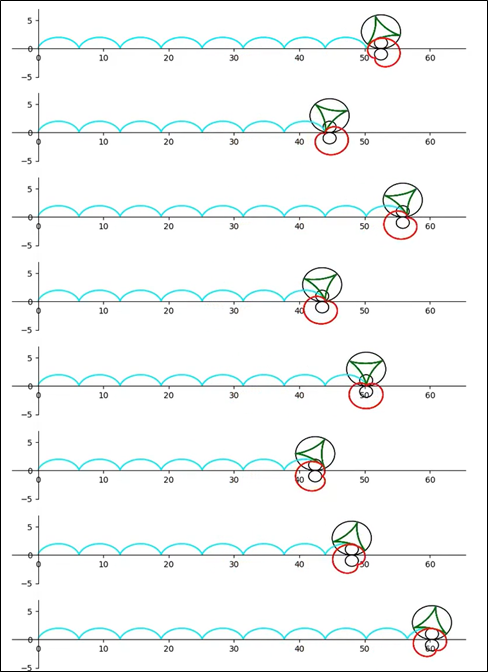} % Reduce the figure size so that it is slightly narrower than the column.
\caption{Structural view of the racing mechanics of a Functional Network of 8 competing neurons which can stop at a preset deadline whether set in time or distance.}
\label{2}
\end{figure}

\begin{figure}[t]
\centering
\includegraphics[width=1\columnwidth]{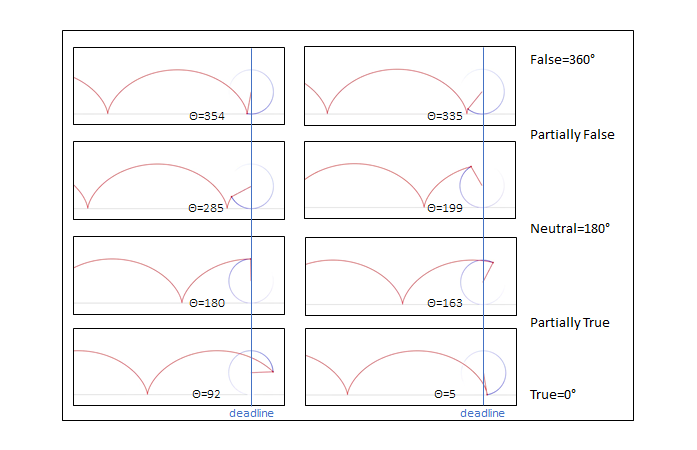} % Reduce the figure size so that it is slightly narrower than the column.
\caption{Value of the angle $\theta$ at the deadline determines the state of the neuron and the output value.}
\label{3}
\end{figure}

\subsection{Functional Neural Network (FNN) and Fuzzy Logic}

The mechanics of the Functional Mirror Network is based on a pool of competing mirror neurons made of neurons in memory (memorial) and a new neuron (request neuron), that move forward until they reach a preset deadline where the state of each individual neuron is determined by the value of the angle $\theta$ (see figure 3) and will qualify the network for a sum state and deliver a prescription to make a decision regarding the request. Figure \ref{2} shows a structural view of the racing mechanics. 

The FNN model is an intention-based decision processing system that delivers an output in the form of a sum of states. The competing neurons are reaching the deadline in different states and the network is deemed to deliver a prescription according to the sum of states at the deadline. Then the output result can trigger a near-natural decision such as buy or sell, yes or no on behalf of the user. In other words, this personal neural network has a human fuzzy logic effect and the idea of extending each personal neural network to other personal networks of similar structure is only logical and the sum of all these networks can help to understand synchronicity in human societies, organized markets and crowd psychology regarding decision making especially these involved in crisis whether social, financial, medical or environmental.

Modelling human fuzzy logic for decision processing means that each neuron has many-valued output. The cycloid movement of the motor core along the timeline ensures that there are 360 values of truth between the 1 or false and the 0 or true, both inclusive. Accordingly, the confidence score of the result will be 1 minus the value of truth e.g. 0\% at 1 (false) and 100\% at 0 (true). This output model delivers a binary response with confidence score. Figure \ref{3} illustrates the correspondence between angle degree in the cycloid movement and output value.

In terms of formulas, we can express the binary response and confidence score for motor core of a node as 
\begin{align*}
&\text{Binary response} = \frac{(v_m t) \% (2\pi r_1)}{2\pi r_1}\\
&\text{Confidence score (in \%)} = (1- \frac{(v_m t) \% (2\pi r_1)}{2\pi r_1}) \cdot 100
\end{align*} 
where $v_m$ will be the velocity of the motor core.

\quad 

Similarly, we can express the binary response and confidence score for sensory core of a node as 
\begin{align*}
&\text{Binary response} = \frac{(v_s t) \% (2\pi r_2)}{2\pi r_2}\\
&\text{Confidence score (in \%)} = (1- \frac{(v_s t) \% (2\pi r_2)}{2\pi r_2}) \cdot 100
\end{align*} 
where $v_s$ will be the velocity of the sensory core.

\quad 

For the FNN, we can define also the concept of group binary response and confidence score of the motor cores and the sensory cores in the network respectively. In terms of formulas, we can express them as follows.
For motor cores of $N$ nodes we have,
\begin{align*}
&\text{Group binary response} = \frac{1}{N}\sum_{i=1}^N\left[\frac{({v_m}_{i} t) \% (2\pi {r_1}_{i})}{2\pi {r_1}_{i}}\right]\\
&\text{Group confidence score (in \%)} =\\
& \quad \left( 1- \frac{1}{N}\sum_{i=1}^N\left[\frac{({v_m}_{i} t) \% (2\pi {r_1}_{i})}{2\pi {r_1}_{i}}\right]\right) \cdot 100
\end{align*} 
where ${v_m}_{i}$ are be the velocity of the $i$-th motor core.
On the other hand, for sensory cores of $N$ nodes we have,
\begin{align*}
&\text{Group binary response} = \frac{1}{N}\sum_{i=1}^N\left[\frac{({v_s}_{i} t) \% (2\pi {r_2}_{i})}{2\pi {r_2}_{i}}\right]\\
&\text{Group confidence score (in \%)} =\\
& \quad \left( 1- \frac{1}{N}\sum_{i=1}^N\left[\frac{({v_s}_{i} t) \% (2\pi {r_2}_{i})}{2\pi {r_2}_{i}}\right]\right) \cdot 100
\end{align*} 
where ${v_s}_{i}$ are be the velocity of the $i$-th sensory core.

\quad 

From the group binary response and confidence score of the motor cores and sensory cores, we can deduce and analyze the expert consensus and group consensus and derive the various additional functions based on these computations, for example, after finding the best performers in terms of the binary response, we can compute the ideal velocities for those low performers to boost in order to achieve the performance of the best performers. We are going to layout the whole framework in the following sections.

In our FNN model, the loss function is based on the binary response of the nodes. The loss function captures the difference between the predicted binary response and the correct decision.
\begin{align*}
\text{Loss function} &= \| y_{predicted} - y\|
\end{align*} 
where $y_{predicted}$ is the predicted binary response, which can be in terms of individual or group, and $y$ is the correct decision.

\subsection{Important algorithms and methods of FNN}

The functional mirror network for decision processing is a cognitive computing design that aims to describe the following mechanisms of the brain in reference to two distinctive groups of algorithms and new methods regarding the executive decision and/or the intuitive decision including calculating the time response of a single neuron, finding the expert consensus in the network, boosting the velocity of certain neurons based on the preset deadline, anticipating the best performer, the best result, the best performance and planning for boosting reward. 

\subsubsection{Modeling executive decision, new methods and algorithms for the motor core:}

\begin{itemize}
\item MOTOR TIME RESPONSE: to compute the average angle $\theta$ (see figure 3) and the group binary response and confidence score of the motor cores in the network according to the rule of the fuzzy logic with preset time deadline;
\item MOTOR DISTANCE RESPONSE: to compute the average angle $\theta$ (see figure 3) and the group binary response and confidence score of the motor cores in the network according to the rule of the fuzzy logic with preset distance deadline;
\item MOTOR NET COMPETE: to select the best performers based on their motor cores performance in the network according to the rule of the fuzzy logic with a preset time deadline. This method can contribute to building an “expert mirror network”; 
\item MOTOR BOOST REQUEST: to boost the velocity of low performers motor core according to the ratio of the best performers. In other words, to compute how much the velocity (in \%) of low performance motor cores should be increased to meet the performance of the best performers.
\item MOTOR BOOST REWARD: to call both the MOTOR NET COMPETE and MOTOR BOOST REQUEST in order to anticipate reward. Based on the new average binary response and confidence score computed in the network for a specific request, this method combines the expert mirror network (best performers per request type) with boost request (increased velocities) to anticipate performance and reward.
\item IDEAL TIME: to predict best results according to a time deadline comparatively to the function of best performers.
\item IDEAL DISTANCE: to predict best results according to a distance deadline comparatively to the function of best performance.
\end{itemize}

\subsubsection{Modeling intuitive decision, new methods and algorithms for the sensory core:}

\begin{itemize}
\item	The SENSORY TIME RESPONSE: to compute the angle $\theta$ (see figure 3) angle and the group binary response and confidence score of the sensory cores in the network according to the rule of the fuzzy logic with preset time deadline;
\item	SENSORY DISTANCE RESPONSE: to compute the angle $\theta$ (see figure 3) angle and the group binary response and confidence score of the sensory cores in the network according to the rule of the fuzzy logic with preset distance deadline;
\item	SENSORY NET COMPETE: to select the best performers based on the sensory cores in the network according to the rule of the fuzzy logic with a preset time deadline;
\item SENSORY BOOST REQUEST: to boost the velocity of low performers of sensory core according to the ratio of the best performers. In other words, to compute how much the velocity (in \%) of low performance motor cores should be increased to meet the performance of the best performers.
\item	SENSORY BOOST REWARD: to call both the SENSORY NET COMPETE and MOTOR BOOST REQUEST to anticipate reward. Based on the new average binary response and confidence score computed in the network for a specific request, this method combines the expert mirror network (best performers per request type) with boost request (increased velocities) to anticipate performance and reward.
\end{itemize}

\section{Future research directions and applications}

\begin{figure}[t]
\centering
\includegraphics[width=1.0\columnwidth]{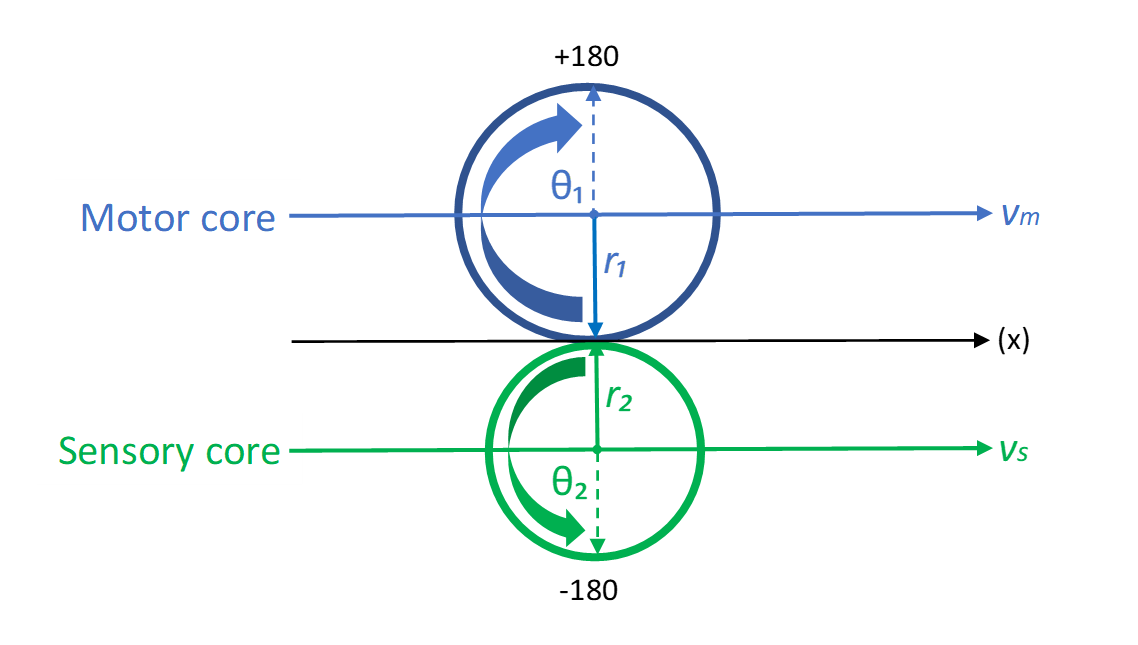} % Reduce the figure size so that it is slightly narrower than the column.
\caption{The velocity and radius of motor core and sensory core can be different.}
\end{figure}

\begin{figure*}[t]
\centering
\includegraphics[width=0.8\textwidth]{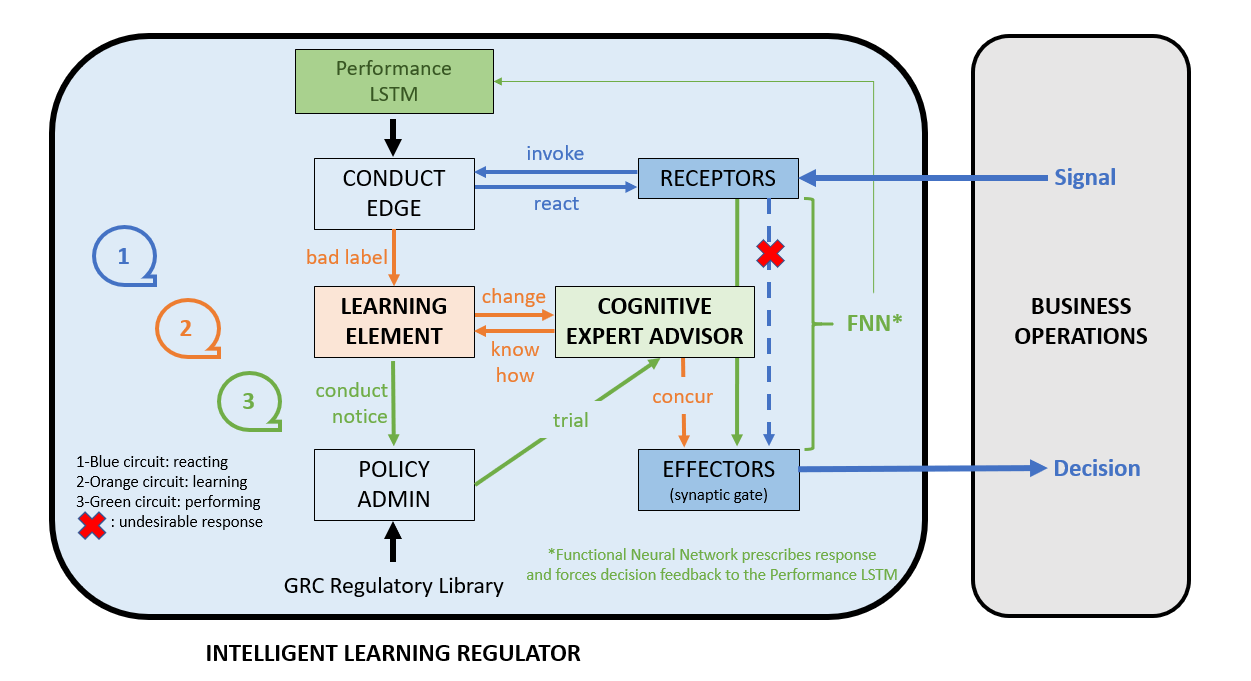} % Reduce the figure size so that it is slightly narrower than the column.
\caption{Intelligent Learning Regulator for business operations supervision.}
\end{figure*}

\subsection{Simulations in Finance} 

One of the key parameters of the FNN is the velocity of the cores of different nodes competing against a preset deadline. In the current configuration, we are assuming that all velocities are set when the nodes are initialized. Based on this initial setup, we have developed some core functions such as NET COMPETE, BOOST REQUEST and BOOST REWARD. But we are keen to extend the concept of velocity to allow acceleration and deceleration of the nodes to simulate scenarios closer to real life such as fast-cycling markets risk management and high frequency trading. This research direction can not only lead to testing new models of decision making for the actors of financial markets, but also enable more accurate simulations in finance and trading that can also be applied to the supply chain in anticipation of the demand.

\subsection{Simulation in Medicine}

In the methodology, we have separated the algorithms and methods of the FNN into two groups, one of them relates to the motor core and the other to the sensory core. It is important to note that although motor core and sensory core are both defined on the same x-axis and share some mathematic properties, they may not be identical as shown in figure 4. Motor core and sensory core are not necessarily synchronized depending on their respective sizes and velocities. That leads us to introduce the concepts of relative synchronicity and true asynchrony between the executive and intuitive parts of a same decision process or between different processes regarding the motor and sensory cores' behaviors. This research direction opens the way to simulation of the learning process(es) whether by accelerating, slowing down, or controlling it with applications in education such as computer-assisted training, enhanced learning, special needs education for autists, or cognitive behavioral therapy (CBT) such as desensitization to control phobias and anxiety, and reconditioning. It could also allow psychiatrists and neuropsychologists to better understand the relationship between disorderly learning and personality disorders, addictions, and criminal misconducts.

\subsection{Learning agent with application in Artificial General Intelligence}

Apart from the direct applications in the fields of finance and medicine, we can also foresee the potential of the FNN in the design of learning agent for Artificial General Intelligence (AGI). AGI is a strong AI that has the capacity to understand or learn to execute an intellectual task like a human being. Because FNN has the capacity to process a decision, prescribe a decision trend while forcing the feedback into a performance memorial, it could act as a cognitive advisor rendering cognitive computing services that can mediate human-machine interactions and enhance significantly the performance of complex business environments such as regulated financial markets. In this sense, AGI can become an intelligent regulator for business supervision as shown in Figure 5. In this design, the receptors are a selection of pre-synaptic neurons running in FNN(s) and the effectors are post-synaptic neurons that are connected to the FNN by synaptic gates for performance optimization. The gating and firing will take advantage of the feedback from the decisions made by the user in relation to the FNN prescriptions, whether correct or not, and from the final effect of the decision on real business operations. The feedback will create a retrograde signaling to affect the gates' subsequent firings. The learning patterns will accumulate in the LSTM performance memorial to serve as a database of the conduct edge. A conduct edge is a subplot using a database of observed conducts or performance memorial (LSTM) to react to an incoming signal captured by the receptors in the business environment. On the other side of the Learning Element, the Policy Administrator is another subplot that can cause the Governance, Risk management and Compliance (GRC) library to trial the conduct notice issued upon reception of a bad label, and make the Cognitive Expert Advisor learn, concur and effect a decision.

\section{Conclusion}

In this paper we have proposed a novel model of Artificial General Intelligence (AGI), the Functional Neural Network (FNN) for modeling of human decision-making processes. The FNN is made of multiple Artificial Mirror Neurons (AMN) racing in the network. Each neuron has the same structure comprising motor core, sensory core and intention wheel and a specific velocity. We have discussed the structure of a simple AMN and its mathematical formulation. We have illustrated the racing mechanism of multiple nodes in the FNN, the group decision process using fuzzy logic and how to transform these conceptual methods into practical methods of simulation and in operations. Finally, we have presented possible future research directions in the fields of finance, education and medicine including the opportunity to design an intelligent learning agent with application in AGI. We believe that FNN promises to transform the way we can compute decision-making and lead to a new generation of AI chips for seamless human-machine interactions.

\subsubsection{Acknowledgment.}
The theory presented in this paper is based on three of our patents, namely, "Method for Informed Decision Making" and "System for Informed Decision Making", Frederic Andre Jumelle, Yu Zhao and Yat Wan Lui, Hong Kong short-term patents grant certificate No. HK30012597A and No. HK30024011A, and "Method and System for Informed Decision Making", Frederic Andre Jumelle, Yu Zhao and Yat Wan Lui, international patent application No. PCT/CN2019/124550.

\bibliography{NIPS_Bib}
\end{document}